\documentclass{article}

\usepackage{arxiv}
\usepackage{enumitem}
\usepackage[utf8]{inputenc} 
\usepackage[T1]{fontenc}    
\usepackage{hyperref}       
\usepackage{url}            
\usepackage{booktabs}       
\usepackage{amsfonts}       
\usepackage{nicefrac}       
\usepackage{microtype}      
\usepackage{lipsum}		
\usepackage{graphicx}
\usepackage{doi}
\usepackage{amsmath}
\usepackage{amsfonts} 
\usepackage{multirow}
\usepackage{tabularray}
\usepackage{float}

\title{Improving Sequence-to-Sequence Models for Abstractive Text Summarization Using Meta Heuristic Approaches}

\author{
    Aditya Saxena \\
    Department of Electronics \& Communication\\
    Bharati Vidyapeeth College of Engineering\\
    Delhi, India \\
    \texttt{adityasaxenaece2@bvp.edu.in} \\
    \And
    Ashutosh Ranjan \\
    Department of Electronics \& Communication\\
    Bharati Vidyapeeth College of Engineering\\
    Delhi, India \\
    \texttt{ashutoshranjanece2@bvp.edu.in} \\
}

\date{}

\renewcommand{\headeright}{}
\renewcommand{\undertitle}{}
\renewcommand{\shorttitle}{\textit{}}

\hypersetup{
pdftitle={IMPROVING SEQUENCE-TO-SEQUENCE MODELS FOR
ABSTRACTIVE TEXT SUMMARIZATION USING META HEURISTIC
APPROACHES},
pdfsubject={CS.ML},
pdfauthor={Aditya Saxena, Ashutosh Ranjan},
pdfkeywords={Natural Language Processing (NLP), Machine Learning (ML), Sequence to Sequence, Encoder-Decoder, LSTM, Transformer Attention, Sequence},
}

\begin{document}
\maketitle

\begin{abstract}
	As human society transitions into the information age, reduction in our attention span is a contingency, and people who spend time reading lengthy news articles are decreasing rapidly and the need for succinct information is higher than ever before. Therefore, it is essential to provide a quick overview of important news by concisely summarizing the top news article and the most intuitive headline. When humans try to make summaries, they extract the essential information from the source and add useful phrases and grammatical annotations from the original extract. Humans have a unique ability to create abstractions. However, automatic summarization is a complicated problem to solve. The use of sequence-to-sequence (seq2seq) models for neural abstractive text summarization has been ascending as far as prevalence. Numerous innovative strategies have been proposed to develop the current seq2seq models further, permitting them to handle different issues like saliency, familiarity, and human lucidness and create excellent synopses. In this article, we aimed toward enhancing the present architectures and models for abstractive text summarization. The modifications have been aimed at fine-tuning hyper-parameters, attempting specific encoder-decoder combinations. We examined many experiments on an extensively used CNN/DailyMail dataset to check the effectiveness of various models.
\end{abstract}


\section{Introduction}
Today, the volume of data and resources has reached an immeasurable size. Other than websites, sources like social media, blogging websites and news forums are increasing the amount of textual content. Not only this, but journals, research papers, novels and stories are also lengthy pieces of documents. People find inconvenience in reading long text, a huge amount of data. Summarization allows users to skip past this inconvenience. Users mostly aim to quickly access information. Summarization allows this by providing a concise and succinct form of the long text thus helping users to save time. Earlier manual summarization was done, but even though the method was effective it was costly and very time taking. With the advent of Automatic Text, Summarization solutions are getting more efficient day by day. There are two ways text summarization can be approached: Abstractive and Extractive. In Extractive text summarization, key phrases and sentences are fetched from the article and consolidated to form the summary. So, in this type of summarization, the content of the summary is entirely from the original text. Abstractive text summarization involves paraphrasing sentences, and new words and sentences can be seen in the summary. Hence, summaries generated by abstractive methods may contain new words, phrases etc. Extractive summarization is based on three tasks: mainly construction of an intermediate representation of input text, assigning scores to the intermediate representation and finally consolidating the summary. The representation of the text can be done in two ways: the first one is a topic representation and the second is an indicator representation. After generating the intermediate representation, a score is assigned to that text to decide how good or accurate the generation is. Finally,  after the above two steps, a complete summarizer system picks the top k most important topics of the text and generates the summary. Although the results provided by extractive methods are great the summaries generated by them are not similar to what humans produce. The reason is that the summaries using this method are based on already present words in the input text, whereas a human-written summary makes use of new words, phrases, and sentences to produce summaries of the text. Hence to tackle this big issue, a study on Abstractive text summarization started. 
\par
Abstractive text summarization is done using the sequence-to-sequence model which was proposed by \cite{Nicolas2007}. Sequence-to-Sequence models consist of an encoder and a decoder. As the name suggests, the encoder encodes the text which means that when input is fed into the model, the encoder is responsible for generating the final state vector which is based on previously fed inputs. After this at each time step decoder is used to predict or generate the summary. Recurrent neural networks successfully convert the text from one form to another in experiments. Machine translation and speech recognition are two examples of such applications. These models are trained on many inputs and anticipated output sequences, and they may then construct output sequences from inputs that have never been presented to the model before. Recently, recurrent neural networks have been used to improve reading comprehension. The models are then taught to remember facts or assertions from the input text. This research is similar to others that employ a neural network to produce news headlines from the same dataset. The primary difference between this and previous work is that instead of employing a recurrent neural network for encoding, they use a simpler attention-based model. In Natural Language Processing, text summarization has become one of the most significant yet complex tasks (NLP). Deep learning’s resurgence has dramatically advanced this field through neural models. A few of the architectures that are prominent for the Abstractive text summarization are RNN-LSTM with coverage mechanism, Pointer Generator Network, and Transformers which can be trained from scratch. Also, we also have an option of using pre-trained models, which is known as transfer learning, in which an already trained model is used for the new dataset that suits a particular use case. This generally produces great results. With every machine learning or deep learning algorithm, optimization plays an important role in achieving the state-of-the-art results. Optimization is the process of achieving the best results for a given dataset. For optimization, multiple algorithms have been produced like standard gradient descent which makes use of slopes to reach the optimal results or Adam Optimizer which is slightly better than SGD. But recently, Bio-inspired algorithms have changed the dynamics of optimization. 
\par
Bio-inspired algorithms excel in addressing complex and combinatorial challenges, with notable examples including Particle Swarm Optimization \cite{Nicolas2007} and Ant Colony Optimization \cite{Mesleh2008}. Among these, the Particle Swarm Optimization stands out as a particularly effective and well-regarded strategy. Although its application has been proven in scenarios such as the Arabic Speaker Recognition System \cite{Harrag2014}, its use in Arabic Text Classification remains unexplored. This research introduces an innovative approach by utilizing Particle Swarm Optimization for feature selection in Abstractive Text Summarization. The key contributions of our study include the development of a Particle Swarm Optimization-based algorithm tailored for feature selection, the introduction of a novel abstractive summarization framework that integrates the Particle Swarm Optimization Algorithm with Support Vector Machines, and the evaluation of our model against prevailing state-of-the-art techniques using a well-established dataset. After the generation of the summary, checking the quality of the summary to find out if it is meaningful or not is very important. There are various metrics available to do so in the field of NLP. The most prominent ones are BLEU and ROUGE. BLEU score, as well as ROUGE score, are n-gram based evaluation metrics which means that it checks how many words of reference summary and machine-generated summary are overlapping. The fundamental difference in both of them is BLEU measures precision which is how many words in the machine-generated summary are present in the human/reference summary whereas ROUGE measures the recall which means how many words in human reference summaries appeared in the machine-generated summaries. 
\par
From the above text, it is evident that both the scores are complementary. Achieving a high BLEU score occurs when numerous words from the system-generated summaries are also found in the reference summaries created by humans. Conversely, a high ROUGE score is obtained when a significant portion of the words in the human-crafted reference summaries are mirrored in the summaries produced by the system. With so much that has been put into Abstractive Text Summarization, this work also proposes a methodology to improve upon the existing architectures. We changed various parameters of existing architectures like the number of layers, the number of cells used, using GRU instead of LSTMs, adding Named Entity Recognition in the summarization pipeline, and many more to improve the ROUGE score (metric for evaluation of most NLP tasks). We proposed these changes majorly to three architectures: LSTM+Attention we call this our base model, the second BASE+coverage mechanism, and finally, Transformer Network. Also, instead of SGD and Adam optimizer, we have proposed the use of Bio-inspired optimization algorithms with some changes in that, so that they are more suitable for text summarization tasks. The objective of this work is to improve on the existing architectures, generate meaningful summaries, tackle the issue of Out-of-vocabulary words, and develop a pipeline so that summaries generated recognized named entities. This article has been fragmented into five sections. In Section 2, we briefly discuss a literature survey on text summarization; this is followed by Section 3, which outlines the proposed methodology of approaching our work. We have discussed and analyzed our results in Section 4, and finally, in Section 5, the article is concluded.

\section{Related Work}
\par
Natural language processing is a tremendously active area of research and development and various approaches have been proposed. Ao, X Wang \cite{Ao2021} used a Microsoft News dataset named PENS dataset. Six user modeling methods were used and compared. Singh \cite{Singh2021} aimed to develop an end-to-end methodology that generates summaries and succinct headlines. The suggested model integrates extractive and abstractive methods through a pipelined approach to provide a succinct summary. The results demonstrated that the proposed approach effectively produces a concise summary and a fitting headline. They used recurrent neural networks \cite{Hayashi2018}, which are based on the machine-translation technique to generate headlines. The Headline generators are composed of an encoder and a decoder and are developed using LSTM. They faced the problem that sometimes the model generated meaningless headlines. A new model was proposed \cite{Li2021} Based on a generative pre-trained model. The proposed model contained only a decoder rather than the encoder-decoder architecture. The experimental results achieved comparable results. However, problems like out-of-vocabulary and the wrong generation of words still existed. A simple approach for news data clustering \cite{Shavrina2021} and the headline generation based on pre-trained language models with few-shot models and minimal fine-tuning was proposed. The presented headline creation method yields 0.596 BLEU and 0.292 ROUGE metrics. In \cite{SinghB2021}, Singh B successfully overcame some of the drawbacks of the transformer, pointer generator, and attention text summarizing architectures, which were supposed to turn circumlocutory news pieces into crisp headlines.. Lopyrev \cite{Lopyrev2015} developed a method for generating headlines from the text of news stories using an encoder-decoder recurrent neural network, LSTM units, and attention, which came out to be one of the first use-cases for the encoder-decoder architecture for the title generation. 
\par
Automatic headline generation for Myanmar News articles using Seq2Seq \cite{Thu2020} with one-hot encoding and describing the comparative analysis result. The errors analysis of a classic headline generation using neural networks and ROGUE evaluation of the headlines generated by machines and actual headlines was performed to attenuate the gap between actual and desired outcomes. RIA and Lenta datasets of Russian news were used to fine-tune the two pre-trained transformer-based models for the news headline generation \cite{Bukhtiyarov2020} and achieved new progressive results. BertSumAbs increased ROUGE on an average by 2.0 points surpassing the previous best score achieved by Phrase-Based Attentional Transformer and CopyNet. Used a sentence extraction method to generate extractive summaries \cite{Mauro2017}. In this method, the sentences are examined for relevancy and rated accordingly using this procedure. The most informative sentences were identified based on sentence scores after similar sentences were clustered together. On the New York Times Annotated Corpus, the Universal Transformer architecture combined with the byte-pair encoding technique \cite{Gavrilov2019} generated a combative ROUGE-L F1 score of 24.84 and the ROUGE-2 score of 13.48. A new network architecture \cite{Vaswani2017}, exclusively depends on attention mechanisms with no recurrence or convolutions. Research into two distinct tasks of machine translation showed these models exceeded their predecessors in quality, ability to parallel process, and shorter training times. Notably, for the WMT 2014 English-to-German translation challenge, the model attained a 28.4 BLEU score, advancing past the best prior achievements by more than 2 BLEU points, even outperforming ensemble methods \cite{Vaswani2017}. The integration of structure learning into graph-based neural models for generating headlines aimed to empower the model to independently learn sentence structure through a data-driven method, utilizing a comprehensive and in-depth network to embed rich relational data among sentences for sentence graph learning \cite{Zhang2020}. 
\par
Hammo et al. \cite{ElHaj2010} developed a method for Arabic text summarization focused on text structure and topic identification, achieving an impressive 92.34\% accuracy. Al-Omour and Al-Taani \cite{AlOmour2014} introduced a hybrid Arabic text summarization approach combining graph-based and statistical methods, which demonstrated promising outcomes. The approach by Ibrahim et al. \cite{Ibrahim2013}, which isolates the most pertinent paragraphs for summarization using cosine similarity, showed its limitations when applied to larger documents. The utilization of the Firefly algorithm for summarizing individual Arabic documents was proposed \cite{AlAbdallah2019}, compared against two evolutionary strategies that involve genetic algorithms and harmony search. This method was evaluated using the EASC Corpus and the ROUGE toolkit, indicating competitive or superior performance compared to the leading methods. Rautray \& Balabantaray \cite{Rautray2015DEPSO} introduced a general-purpose document summarizer using a particle swarm optimization algorithm, focusing on content coverage and redundancy as the primary summary features. A similar PSO-based summarizer was later described \cite{Rautray2015SentenceFeatures}, which, instead of utilizing sentence weights, directly incorporated textual features into its model. This summarizer also assessed summary attributes such as content coverage, readability, and length. Gupta, Sharaff, and Nagwani’s work \cite{Gupta2024} introduces the Sim-TLBO framework, significantly enhancing biomedical text summarization through optimized sentence selection, evidenced by notable advancements in ROUGE scores and readability. Although numerous optimization algorithms \cite{ModiriDelshad2016, Kaboli2016Forecasting, Rafieerad2017, Kaboli2017Rainfall, Sebtahmadi2017, Kaboli2017Energy}
 have been explored in literature, the development and refinement of text summarization techniques continue.

\section{Methodology}
\subsection{Data Preproccessing}
\subsubsection{Data Cleaning}
A very common dataset on which most of the models are trained and are widely accepted is CNN/Dailymail dataset. The dataset contains extracts from the news articles of CNN and the Dailymail news channel. There are approximately 90k documents. The document has an article and its summary, which we use to train our model. 
The stories obtained were downright messy and unclean text with many redundant characters and symbols. So we had to clean the data and extract summaries from the story articles.

\begin{enumerate}[label=\Alph*.] 
    \item Expanding the contraction words:\\
    In our normal conversations, we tend to use words like can’t, don’t, I've, etc. which are not understandable to machines. These words are called contraction words where we combine two words to shorten them. Since machines can’t understand these words we expanded these as a part of our cleaning process.
    
    \item Removing unwanted symbols:\\
    The article in every document that we used contained Token ‘(CNN)’ at the start of it which doesn’t make any sense. So we had to remove it, also symbols like ‘\$ \%\^{}\&\* \#’ do not have any meaning to the text so these were also removed.

\end{enumerate}

\subsubsection{Text Analysis}

\begin{enumerate}[label=\Alph*.] 
    \item Article Length: Figure~\ref{fig:article_length} describes the univariate distribution plot for the article lengths basically it gives you an idea of what is the most common word length, or what is the range of length that encompasses most of the articles 

    \begin{figure}[h]
        \centering
        \includegraphics*[width=4.32in, height=1.70in]{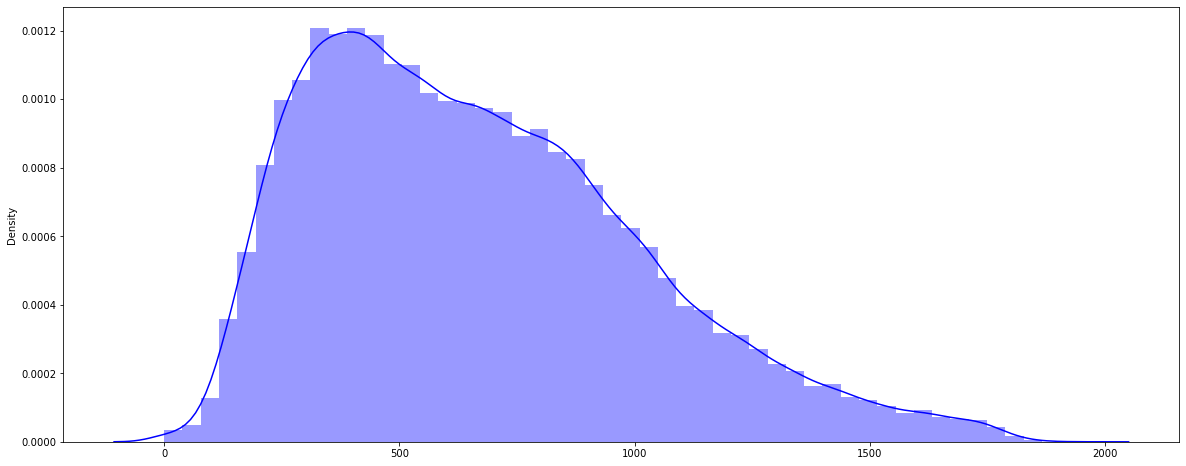} 
        \caption{Article Length Distribution}
        \label{fig:article_length}
    \end{figure}
    
    \item Summary Lengths: Figure~\ref{fig:summary_length} depicts the distribution of summary length 

    \begin{figure}[h]
        \centering
        \includegraphics*[width=4.32in, height=1.70in]{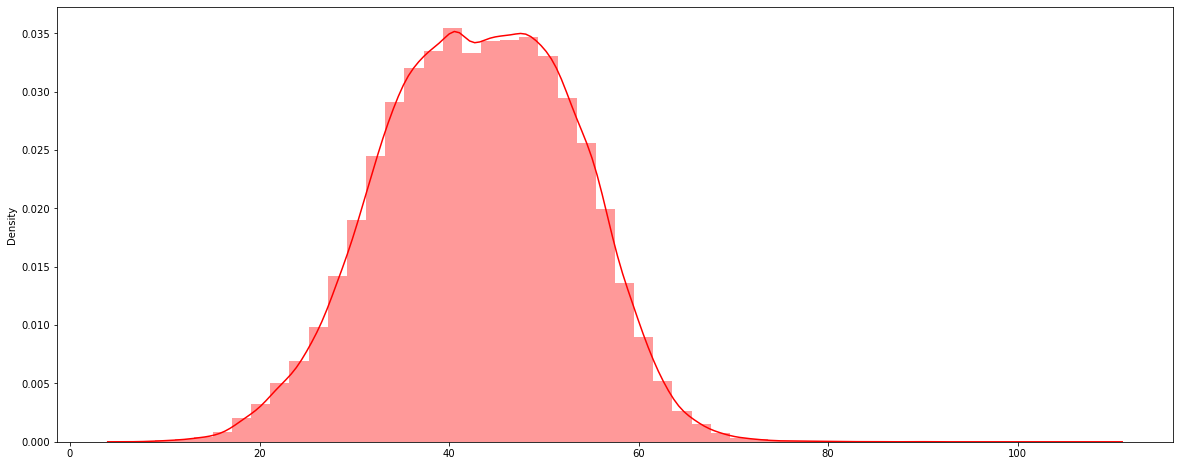} 
        \caption{Summary Length Distribution}
        \label{fig:summary_length}
    \end{figure}
    Most of the summaries have a length of 35-55 words,

    \item Word cloud for summary words: Word clouds are visualizations generated to get a high-level idea of the frequency of a word in text data. It was done using the python WordCloud library. The size of each word is an indicator of how frequently that word has occurred in our dataset or vocabulary. The larger the size of the text, the greater its frequency, as we can see in Fig3 that words like say, will, new are comparatively more significant than the other words. This is because these are very common verbs, and hence their frequency is higher. As CNN articles are American news-based, words like 'American,' 'New York,' 'The United States' hold significant importance in the cloud.

    \begin{figure}[h]
        \centering
        \includegraphics*[width=3.45in, height=3.17in]{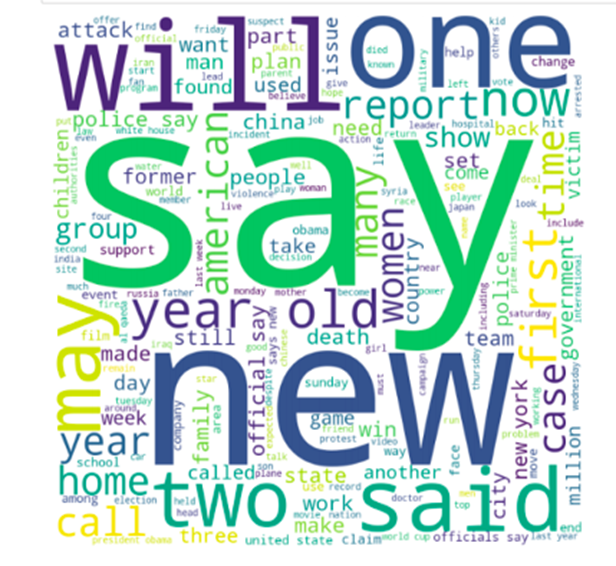} 
        \caption{Word Cloud}
        \label{fig:word_cloud}
    \end{figure}

    \item Named Entities Tagging: Named Entity are the names of the organization, personality , places etc. Now for a summarizer or, for that fact, any NLP model, it is challenging to understand named entities, so to avoid this, we include NER in our model’s pipeline, which means whenever data is passed through the model, first-named entities are tagged so that model can learn that and predict the same while decoding the process.

\end{enumerate}

\subsubsection{Data Pre-processing}

\begin{enumerate}[label=\Alph*.] 
    \item Handling OOV Words: \\
    A model cannot know every word in the world. Hence we need to find a way to tackle vocabulary words. So to handle this issue we replace the word not present in the training set by ‘unk’ which stands for unknown , all the words not in the training vocab will be replaced by this token .During the testing phase , the untrained vocab is provided as ‘unk’ to the model. Now if ‘unk’ is not trained during the training phase it would reduce the capability of the model vastly. So, we need to include the ‘unk’ token during the training phase. For that, we consider rare words in the training vocab as ‘unk.’ Also, to reduce out of vocabulary words, we used a more significant word embedding so that significantly fewer out vocabulary words are encountered.

    \item Tokenizing \& Padding: \\
    We used Keras tokenizer to tokenize words. After initializing the tokenizer, it is fitted on the training data. Then it will also be used while testing. For maintaining a standard, we needed to set the maxlen for both article and summary. If during training, any text has a length exceeding maxlen, the exceeding part of the sequence will be padded with 0 and will not have any significant role in training.

    \item Word Embeddings: \\
    Machines do not understand words, text values. They understand data in terms of matrix, vectors, or numbers. So for that, we need to convert our input text into a numerical representation. Almost every NLP use case is done by creating word embeddings. It is a technique where single words are interpreted as real-valued vectors in a predefined vector space. Every word in the vocabulary is mapped to a predetermined vector space whose size is the number of words*predetermined dimensions. We used pre-trained GloVe word embeddings of 300 dimensions.

\end{enumerate}

\subsection{Implementation}
\subsubsection{Optimization Algorithms}
Whenever a Deep Learning based task is performed optimization algorithms play a key role in determining how efficient and accurate your model will be. Orthodox algorithms like SGD and Adam in the past have performed really well but bio-inspired optimization algorithms are taking the center stage. We have mentioned some bio-inspired algorithms which we have combined with our models. The rising popularity of meta-heuristic optimization algorithms in engineering disciplines can be attributed to their foundation on straightforward principles, ease of implementation, and independence from gradient data, allowing them to circumvent local optima effectively. These algorithms are versatile, suitable for a broad spectrum of challenges across various fields. Numerous academics and researchers have pioneered the development of diverse metaheuristic strategies to tackle intricate and unresolved optimization issues, including but not limited to Particle Swarm Optimization, Ant Colony Optimization, and Whale Optimization Algorithm

\begin{figure}[h]
    \centering
    \includegraphics*[width=2.63in, height=1.86in]{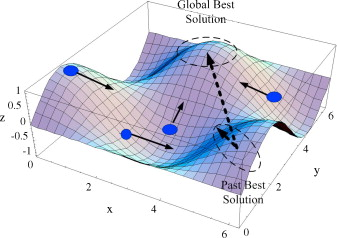} 
    \caption{Particle Swarm Optimization Algorithm}
    \label{fig:pso}
\end{figure}

The Particle Swarm Optimization (PSO) algorithm, drawing inspiration from the collective behavior observed in bird flocks or fish schools, is a method based on the stochastic optimization of a population. Illustrated in Fig~\ref{fig:pso}, the PSO's mechanics involve numerous individual particles, each representing a possible solution navigating the search space with velocity, guided by both their individual and collective experiences. As the particles adjust their paths over time, guided by specific goals or criteria, they all move closer to the best solution together, using rules based on experience and intuition. PSO stands out from other similar algorithms because it can remember past positions, which helps it work faster. It's also good at exploring many options at once, which is perfect for solving problems that involve lots of different factors. This makes PSO particularly well-suited for tackling complicated issues in new and developing areas of technology.

\begin{figure}[h]
    \centering
    \includegraphics*[width=2.80in, height=1.83in]{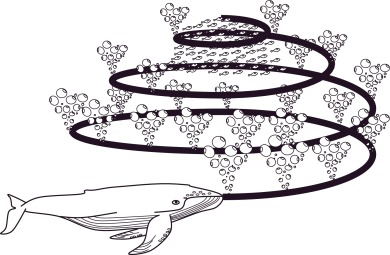} 
    \caption{Whale Optimization Algorithm}
    \label{fig:woa}
\end{figure}

The Whale Optimization Algorithm (WOA) draws its inspiration from the hunting techniques of humpback whales, specifically their unique method known as bubble-net feeding. This method, which is the foundation of the WOA, is shown in Fig~\ref{fig:woa} and involves humpback whales creating spirals or nets of bubbles to trap their food, such as krill or small fish, near the surface of the water. These whales circle their prey, releasing bubbles to form a barrier in either a circular shape or resembling the number 9. This specialized feeding behavior, unique to humpback whales, captures prey within a bubble column, allowing the whales to feed efficiently. The WOA translates this natural behavior into a computational process, aiming to solve optimization problems by adopting the strategic, spiral movement of whales as they close in on their prey, using either a random or the best search path to find the most effective solution.

\begin{figure}[h]
    \centering
    \includegraphics*[width=2.86in, height=1.57in]{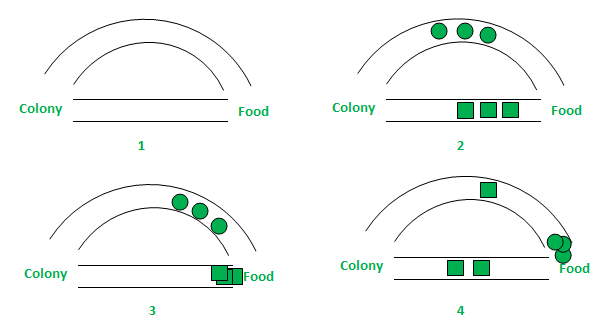} 
    \caption{Ant Colony Optimization Algorithm}
    \label{fig:acoa}
\end{figure}

Introduced by Marco Dorigo in his doctoral thesis during the 1990s, the Ant Colony Optimization (ACO) algorithm (see Fig~\ref{fig:acoa}) draws its inspiration from the natural foraging patterns of ants, as they navigate between their colony and food sources. Initially conceptualized to address the challenging traveling salesman problem, its application has since expanded to tackle a variety of complex optimization issues.

Ants, known for their social structure, inhabit colonies and exhibit collective behavior driven by the pursuit of nourishment. Their foraging process involves a random but purposeful exploration of the surrounding area. As an ant moves, it leaves behind a pheromone trail, a chemical marker that guides other ants to potential food sources. The strength of the pheromone signal, which reflects the quantity and quality of the food, influences the path selection of fellow colony members. Consequently, routes with higher pheromone concentrations attract more ants, leading to a reinforcing cycle where the path with the most traffic becomes the most attractive. This iterative process of exploration and pheromone deposition enables the colony to efficiently discover and exploit food resources.

\subsubsection{Baseline Model + Coverage Mechanism}
The major issue with the basic sequence to sequence models is repeating of words in the generated summary, especially when there are multiple sentences in it. In this coverage mechanism we use a coverage vector c\textsuperscript{t} that is the sum of the attention distributions over all previous decoder timestamps.

\begin{equation}
c^t = \sum_{t'=0}^{t-1} a^{t'}
\end{equation}

Intuitively, a distribution over the source document words addresses the level of coverage that those words have received from the attention mechanism up to this point. c0 is a zero vector because, on the first timestamp, none of the source documents has been covered. Also, a coverage loss is added to penalize repeatedly attending to the exact locations.

\begin{equation}
\mathit{covloss} = \sum_i \mathit{min}(a_i^t, c_i^t)
\end{equation}

We tried different numbers of hidden units to find out the optimal setting. We finally used 256 hidden units. We tried different values for dropout regularization, the number of GRU layers. In both LSTM and GRU units, GRU performed slightly better.

\subsubsection{Transformer Network}
The transformer network is the next step taken in encoder-decoder models; this model is different from vanilla encoder-decoder architecture because it does not implement any Recurrent Network(GRU, LSTM). One of the best ways to implement sequence to sequence until now was Recurrent Networks, but studies showed that models only with attention mechanisms prove to be more robust and effective for many sequential tasks.

\begin{figure}[H]
    \centering
    \includegraphics*[width=3.72in, height=4.5in]{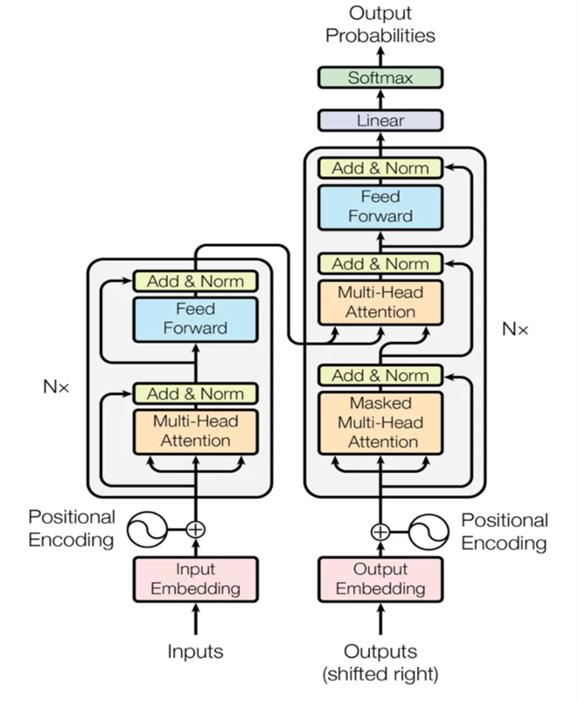} 
    \caption{The Transformer - Model Architecture}
    \label{fig:transformer}
\end{figure}

Fig~\ref{fig:transformer}, shows the original architecture of the Transformer model presented in the paper \cite{Vaswani2017}. It consists of two parts with an encoder on the left and a decoder on the right. Both are composed of multiple blocks and can be replicated to make a complete model. One of the significant problems that were intended to be solved by this was the slowness of the original Encoder-Decoder model. In the original Encoder-Decoder architecture with Attention, input was fed in a sequential manner which does not allow the model to perform parallelization. In contrast, in the Transformer network, input is fed entirely in one single instance rather than sequentially, thus allowing parallelization and speeding up the training process. Since we are feeding a complete sequence at a single instance and Recurrent networks are not present to remember how sequences are fed into the model, it was required to allow a relative position to the sequence. The positions are added to the embeddings of each word. Now, The main essence of the Transformer network is the Self-Attention layer.\cite{Vaswani2017} The primary function of the Self-Attention layer is to capture the context of how relevant a word is w.r.t other words in that particular sequence. However, the problem with having only one self-attention vector gives more value to words currently under processing, so to tackle that, multiple vectors were considered. In the original paper, the number chosen was 8. We tried different numbers and arrived at our final result of choosing 10. This is why there is a block called Multi-Head Attention because we are using multiple blocks of Self-Attention. After Multi-Head Attention generates an attention vector, it is passed through a feed-forward neural network to transform the shape of attention vectors into a form acceptable by the next encoder layer or decoder layer. Now, in the original transformer, there were 6 Encoders. We tried changing it so that it could fit our dataset much better. Now comes the decoder block, the components of which are almost the same with one exception: instead of Multi-Head Attention, it has Masked Multi-Head Attention\cite{ElHaj2010}. This is done because while the training process in transformer network decoding is done using the Teacher Forcing method, now since we are feeding the original summary to the decoder for training purposes, we do not want the model to predict the next word using the original summary instead we want it to learn using previous results only that is why we need to mask it. After this, it is passed to the feed-forward neural network, then to the linear block, and finally to a softmax layer to convert the results into a probability distribution, which is human interpretable. We changed and tested multiple values for multiple hyper-parameters to make the model more robust for our use case.

\subsubsection{Pointer Generator Network}

\begin{figure}[H]
    \centering
    \includegraphics*[width=3.83in, height=1.97in]{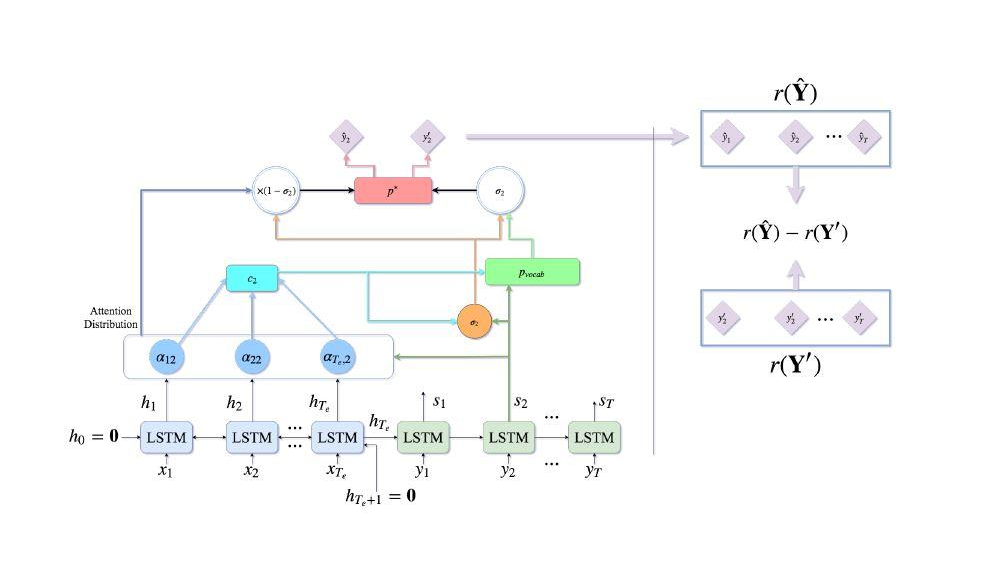} 
    \caption{Pointer Generator Network - Model Architecture}
    \label{fig:pointer_gen}
\end{figure}

In each decoder time-step, as shown in Fig~\ref{fig:pointer_gen}, the generating network produces a vector that modulates content-based attention weights over inputs. These weights are calculated by taking softmax operation with dictionary size equal to that of the length of the input sequence.In the pointer network, these attention weights/masks are not further used to calculate the context vector for the next time step. These weights are considered as pointers to the input sequence. Input time-step having the highest weight is considered as output for that decoder time-step.

\begin{equation}
u_i^j=v^T \tanh \tanh \left(W_1 e_j+W_2 d_i\right) \quad j \in(1, \ldots, n)
\end{equation}

\begin{equation}
p\left(C_1, \ldots, C_{i-1}, P\right)=\mathit{softmax}\left(u^i\right)
\label{eq:softmax}
\end{equation}

From Equation~\ref{eq:softmax}, it can be seen that softmax operation over “u” is not further used to calculate context vectors to feed as information to the current decoder step. The output of softmax operation points to the input token having the maximum value. Consider the output of the first step of the decoder step is “1”. Then for the next time-step corresponding input token representation of input [X1, Y1] along with the decoder is hidden state representation of the previous time step is fed to the network to calculate the hidden state representation of the current time step. The output of the current step is “4”, hence [X4, Y4] goes to the input for the next step. It should be understood that a simple RNN sequence-to-sequence model could have solved this by training to point at the input target indexes directly. However, at inference, this solution does not respect the constraint that the outputs map back to the input indices exactly. Without the restrictions, the predictions are bound to become blurry over more extended sequences.

\subsubsection{Evaluation Metric}

\textbf{ROUGE Metric:}

ROUGE is an acronym that stands for Recall Oriented Understudy for Gisting Evaluation\cite{AlAbdallah2019}. Rouge isn’t a single metric rather it is a set of metrics. For our evaluation, we used three Rouge scores: ROUGE-1, ROUGE-2, ROUGE-L.

\begin{enumerate}[label=\Alph*.] 
    \item \textbf{ROUGE-1:} ROUGE-1 measures the number of matching unigrams between the summary generated by our model and the reference summary. An unigram means a single word, so what ROUGE-1 does is measure how many words in the generated summary and reference summary are the same.

    \item \textbf{ROUGE-2:} ROUGE-2 is similar to ROUGE-1 with a difference that instead of finding out matching unigrams it uses bigrams
    
    \item \textbf{ROUGE-L:} ROUGE-L calculates the longest subsequence which matches with the reference summary. The longer the matched subsequence the higher similarity is there between generated summary and reference summary

\end{enumerate}

\section{Results}
Over recent years, the CNN/Daily Mail dataset has emerged as a widely recognized benchmark for assessing the capabilities of various summarization models, particularly those designed to produce multi-sentence summaries from longer documents. In our experiments, we systematically tested the effects of 3 models: Bi-Directional LSTM with Attention Mechanism, Coverage Mechanism, Transformer Network with 3 bio-inspired algorithms which are Whale Optimization, Particle Swarm and Ant Colony. Our experimental results are shown in Table~\ref{tbl:performance_table}. Although, Bi-directional LSTM with Attention has shown great promise but results even after combining it with different decoder algorithms were not improving. Thus we tried a coverage mechanism that reduced the repetition of words and increased the ROUGE score. But the best performer out of all the algorithms that we experimented with and tweaked was Transformer with Particle Swarm Optimization algorithm. 
\begin{table}[h]
\centering
\caption{Performance evaluation of obtained results}
\label{tbl:performance_table} 
\begin{tblr}{
  cell{2}{1} = {r=3}{},
  cell{5}{1} = {r=3}{},
  cell{8}{1} = {r=3}{},
  vlines,
  hline{1-2,5,8,11} = {-}{},
  hline{3-4,6-7,9-10} = {2-5}{},
}
\textbf{Algorithm} & \textbf{Optimization Algorithm} & \textbf{Rouge-1} & \textbf{Rouge-2} & \textbf{Rouge-L} \\
Coverage Mechanism & Whale Optimization              & 29.4             & 11               & 21               \\
                   & Particle Swarm                  & 29               & 10.9             & 20.7             \\
                   & Ant Colony                      & 27               & 7                & 18.2             \\
Transformer        & Whale     Optimization          & 38               & 15.7             & 25.6             \\
                   & \textbf{Particle Swarm}         & \textbf{41}      & \textbf{17.6}    & \textbf{37.9}    \\
                   & Ant Colony                      & 35               & 14.01            & 23.12            \\
Pointer Generator  & Whale Optimization              & 35.39            & 14.25            & 30.88            \\
                   & Particle Swarm                  & 33.1             & 12.21            & 29.68            \\
                   & Ant Colony                      & 28               & 11.24            & 23.2             
\end{tblr}
\end{table}

Transformer with 8 encoders and 8 decoders worked best for summarization tasks with Particle Swarm Optimization Algorithm. Also, the number layers in the feed-forward network layer were used as a hyperparameter and were used for optimizing the results. It was tested for 4, 6, 8, 10, 16, and 32 layers, 6 layers gave the best results. The number of Self-Attention vectors was also treated as a hyper-parameter which was set at 10 for optimized results The bar graph below shows the comparison of our best model results with the results of the other researches we used as reference. The results obtained by us were better than 2 of them but poor than the third one.

\begin{figure}[H]
    \centering
    \begin{minipage}{.5\textwidth}
        \centering
        \includegraphics[width=3.26in,height=1.99in]{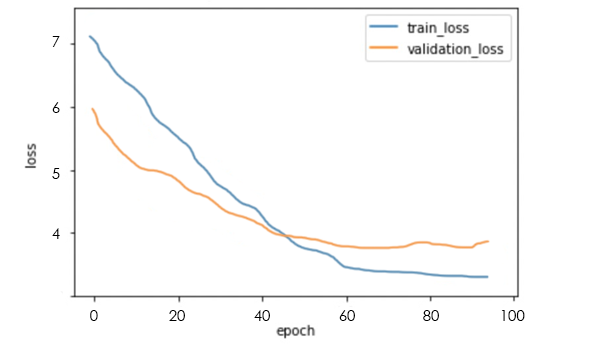}
        \caption{Transformer Network (Whale Optimization)}
        \label{fig:trns_woa}
    \end{minipage}%
    \begin{minipage}{.5\textwidth}
        \centering
        \includegraphics[width=3.26in,height=1.99in]{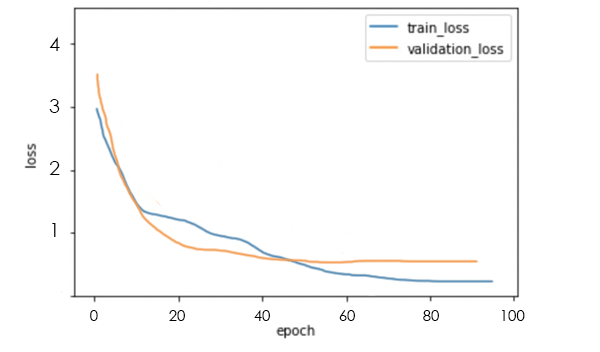}
        \caption{Transformer Network (Particle Swarm)}
        \label{fig:trns_pso}
    \end{minipage}
\end{figure}

\begin{figure}[H]
    \centering
    \begin{minipage}{.5\textwidth}
        \centering
        \includegraphics[width=3.26in,height=1.99in]{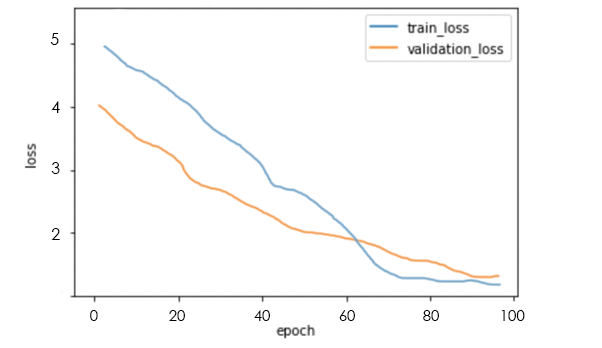}
        \caption{Transformer Network (Ant Colony Optimization)}
        \label{fig:trns_aco}
    \end{minipage}%
    \begin{minipage}{.5\textwidth}
        \centering
        \includegraphics[width=3.26in,height=1.99in]{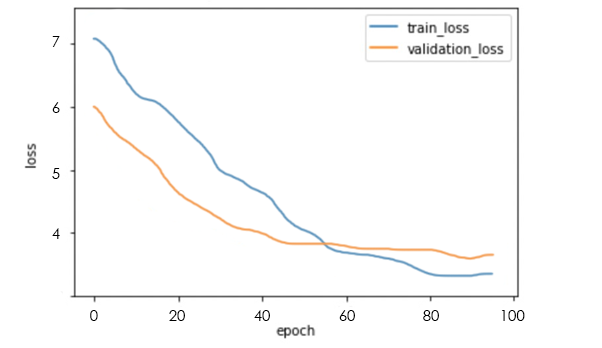}
        \caption{Pointer Generator (Whale Optimization)}
        \label{fig:pointer_woa}
    \end{minipage}
\end{figure}

\begin{figure}[H]
    \centering
    \begin{minipage}{.5\textwidth}
        \centering
        \includegraphics[width=3.26in,height=1.99in]{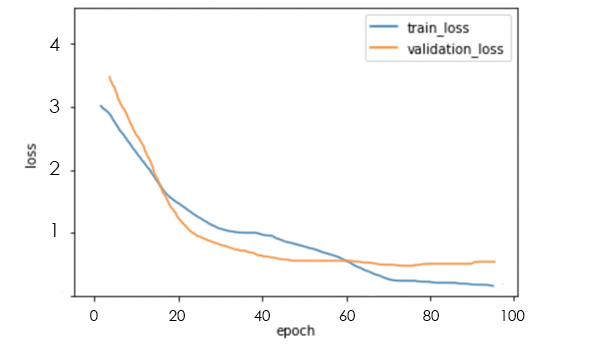}
        \caption{Pointer Generator (Particle Optimization)}
        \label{fig:pointer_pso}
    \end{minipage}%
    \begin{minipage}{.5\textwidth}
        \centering
        \includegraphics[width=3.26in,height=1.99in]{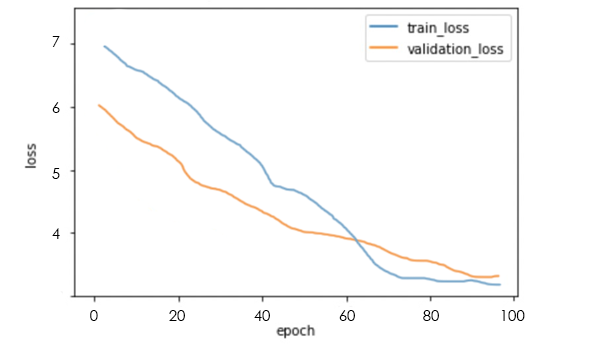}
        \caption{Pointer Generator (Ant Colony Optimization)}
        \label{fig:pointer_aco}
    \end{minipage}
\end{figure}

The loss curve of training and validation for Transformer with 3 optimization algorithms can be seen in Fig~\ref{fig:trns_woa}, Fig~\ref{fig:trns_pso}, Fig~\ref{fig:trns_aco} respectively. We can observe that the results obtained by the loss curve are also better for the Transformer network only. Similarly pointer generators loss curve and validation curve can be seen in Fig~\ref{fig:pointer_woa}, Fig~\ref{fig:pointer_pso} and Fig~\ref{fig:pointer_aco}.

\begin{figure}[H]
    \centering
    \includegraphics*[width=5.14in, height=1.10in]{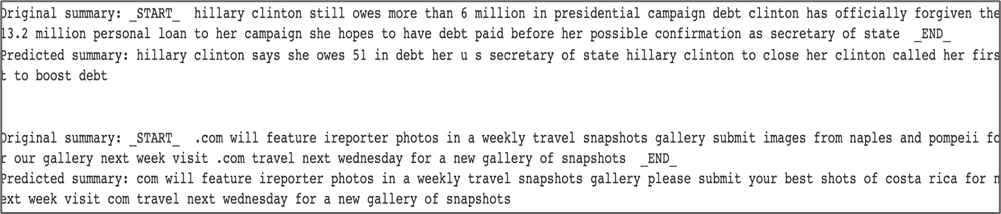} 
    \caption{Summary by Bidirectional LSTM with Attention and Coverage Mechanism}
    \label{fig:result_lstm}
\end{figure}

\begin{figure}[H]
    \centering
    \includegraphics*[width=5.14in, height=1.10in]{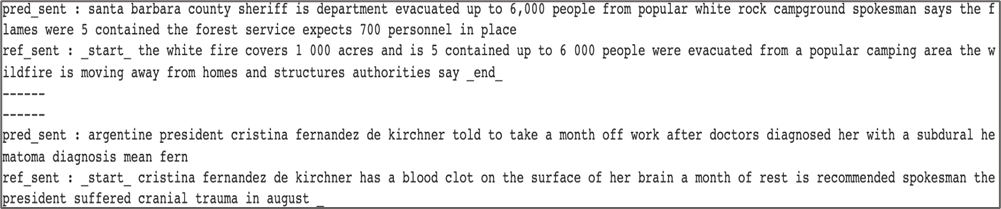} 
    \caption{Summary by Transformer Model}
    \label{fig:result_trns}
\end{figure}

Summaries generated by both the Pointer Generator and transformer network are shown in Fig~\ref{fig:result_lstm} and Fig~\ref{fig:result_trns} respectively. The summaries for coverage mechanism are cherry-picked whereas summaries of the Transformer network were randomly picked as most of the summaries generated by it were of good quality.

\section{Conclusion}
We have tried to improve the performance of existing models for text summarization like transformer coverage, etc. This work primarily focuses on improving the existing architecture to generate summaries, recognize names of entities in an article, and handle OOV words. The results obtained by making changes to architecture improved the results compared to the existing research in the field. Further, training on even larger datasets and experimenting with other architectures will improve the summary's quality. Although recent advancements in sequence to sequence models have improved text summarization, many problems are still at large. Metrics like BLEU score , BERT score , ROUGE score can be used in the training process for reward based training. Hence , with the help of RL based training current models can be improved further. Sampling based approach can be used for the decoding process in place of already widely used beam search as this technique has shown good accuracy as well as diversification in generated text when used for the other text generation based tasks of NLP. The current datasets like CNN/Dailymail have articles written by journalists which already in essence appears to be summary. Because this model gets biased towards extractive rather than abstractive. More generalized dataset of news corpora would be a big step in unlocking the full potential of Abstractive Text Summarisation .Most of the existing evaluation scores like BERTscore , BLEU score are not a very robust and effective way to analysis summarization task reason being the features they take in consideration for evaluating are very limited and not at all comprehensive.Thus a new metric which has more robust approach , which evaluates summary by considering multiple features like the meaning , fluency , if the facts in the summary are correct or not, etc would be more apt.

\end{document}